\documentclass[conference]{IEEEtran}

\RequirePackage{ifpdf}

\ifpdf
  \pdfoutput=1\relax
  \usepackage{graphicx}
  \DeclareGraphicsExtensions{.pdf,.png,.jpg,.jpeg}
\else
  \usepackage{graphicx}
  \DeclareGraphicsExtensions{.eps}
\fi

\usepackage{cite}
\usepackage{amsmath,amssymb,amsfonts}
\usepackage{algorithmic}
\usepackage{textcomp}
\usepackage[dvipsnames]{xcolor}

\usepackage{array}
\usepackage{url}
\usepackage[colorlinks=true,linkcolor=red,urlcolor=blue,pdfstartview=FitH,bookmarksopen=true, citecolor=green]{hyperref}

\usepackage{caption}
\usepackage{listings}

\def\BibTeX{{\rm B\kern-.05em{\sc i\kern-.025em b}\kern-.08em T\kern-.1667em\lower.7ex\hbox{E}\kern-.125emX}}

\begin{document}

\title{Discovering Features in Sr$_{14}$Cu$_{24}$O$_{41}$ Neutron Single Crystal Diffraction Data by Cluster Analysis}

\author{

\IEEEauthorblockN{1\textsuperscript{st} Yawei Hui}
\IEEEauthorblockA{\textit{\small Computer Science and Mathematics Division} \\
\textit{\small Oak Ridge National Laboratory}\\
\small Oak Ridge, TN 37831 \\
\href{mailto:huiy@ornl.gov}{huiy@ornl.gov}
}

\and

\IEEEauthorblockN{2\textsuperscript{nd} Yaohua Liu}
\IEEEauthorblockA{\textit{\small Neutron Scattering Division} \\
\textit{\small Oak Ridge National Laboratory}\\
\small Oak Ridge, TN 37831 \\
\href{mailto:liuyh@ornl.gov}{liuyh@ornl.gov}
}

\and

\IEEEauthorblockN{3\textsuperscript{rd} Byung Hoon Park}
\IEEEauthorblockA{\textit{\small Computer Science and Mathematics Division} \\
\textit{\small Oak Ridge National Laboratory}\\
\small Oak Ridge, TN 37831 \\
\href{mailto:parkbh@ornl.gov}{parkbh@ornl.gov}
}
}

\maketitle

\begin{abstract}
To address the SMC'18 data challenge, ``Discovering Features in Sr$_{14}$Cu$_{24}$O$_{41}$'', we have used the clustering algorithm ``DBSCAN'' to separate the diffuse scattering features from the Bragg peaks, which takes into account both spatial and photometric information in the dataset during in the clustering process. We find that, in additional to highly localized Bragg peaks, there exists broad diffuse scattering patterns consisting of distinguishable geometries. Besides these two distinctive features, we also identify a third distinguishable feature submerged in the low signal-to-noise region in the reciprocal space, whose origin remains an open question.
\end{abstract}


\section{Introduction}
\label{sec:Introduction}

Neutron single crystal diffraction is a powerful way to examine the atomic structure of scientifically and/or technologically interesting materials. The particular material represented in this dataset is from an aperiodic crystal Sr$_{14}$Cu$_{24}$O$_{41}$ (known informally as the telephone number compound)~\cite{b1, b2}, a semi-conducting solid with industrial application potential for optoelectronic and thermoelectric devices. Sr$_{14}$Cu$_{24}$O$_{41}$ has two intertwined structures. One is a Cu$_2$O$_3$ two-leg ladders and the other is CuO$_2$ chains. The ladder and chain structures are incommensurate, which means there is not a simple ratio of the number of chain links to the number of ladder rungs.

Discovered as a byproduct of cuprate synthesis and examined since the 1980s with x-ray, electron, and neutron scattering, as well as Raman and other microscopic methods, the compound structure is still not fully understood. This compound displays rich physical phenomena under small structural disturbance.  Its intertwined-lattice structure gives rise to charge-density wave or anti-ferromagnetism in some circumstances,  and superconductivity in other cases. Single crystal diffuse scattering provides a path to determine the local structure and can potentially provide insight into how the diversified physical properties are related to details of the micro-structure. The single crystal diffuse data can be information-rich but very complex. To date, it is still very challenging to analyze single crystal diffuse scattering for general cases. The goal of the current work is to investigate approaches to extract distinguishable diffuse scattering features for the future data analysis\footnotemark. 

\footnotetext{VIDEO LINK: \url{https://www.youtube.com/watch?v=djx1fA97tO4&t=157s}}

\section{Data Reduction and Analysis}
\label{sec:DataReduction}

\subsection{Data set at a Glance}
\label{sec:Dataset}

\begin{figure}[!h]
	\includegraphics[width=0.48\textwidth]{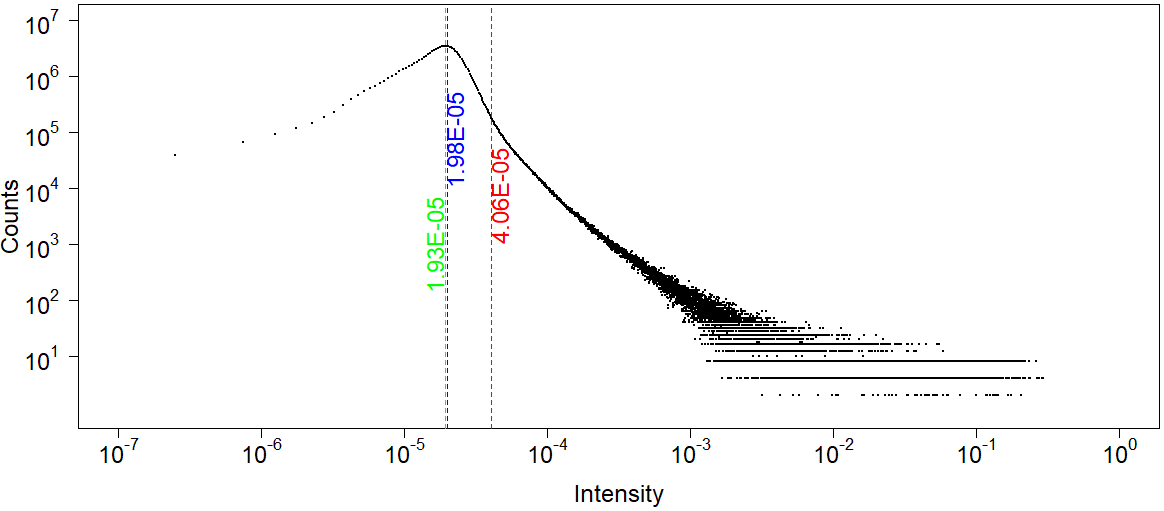}
	\caption{Intensity profile plotted as a histogram, with several characteristic positions: the mean value \textcolor{red}{VMEAN}, the median value \textcolor{blue}{VMEDIAN} and the maximum value\textcolor{green}{HMAX}}
	\label{fig:Histogram}
\end{figure}

The data set provided in this challenge can be represented by a 3D cube in the reciprocal space along three Cartesian coordinate axes of $Q_1\in[-10, 10]$, $Q_2\in[-10, 10]$ and $Q_3\in[-25, 25]$. A regular grid has been set up so that there are 501 evenly spaced grid points along each axis. The original data set thus possesses close to 126 millions data points. At each grid point there is an intensity value that is the 3D spatial Fourier transformation of the neutron-sample interaction potential deducted from proper data reduction process such as~\cite{b3}. To acquire the first order statistics on the data set, we calculate the mean (\textcolor{red}{VMEAN}) and median (\textcolor{blue}{VMEDIAN}) values in the intensity space and plot the intensity distribution in~\autoref{fig:Histogram},, taking a linear binning scheme with total number of bins of $10^6$. The position (in intensity) of the maximum binned counts is marked as \textcolor{green}{HMAX}. These statistical quantities are used to determine some of the initial parameters for the DBSCAN algorithm, which is introduced next.

\subsection{Introduction to DBSCAN}
\label{sec:DBSCAN}

A DBSCAN (Density-based spatial clustering of applications with noise)~\cite{b4} application takes two parameters in the clustering procedure: ``$\epsilon$'' -- the maximum distance between two points for them to reside in the same neighborhood; and ``\textit{minPts}'' -- the minimum number of points required to form a dense region. The distance between two points is usually defined in the Euclidean metric. For the neutron dataset in our work, coordinates of data points are taken as the uniformly distributed voxel indices which scale linearly with the physical positions of data points in the reciprocal space. For a Cartesian coordinate system, if considering only the smallest ``$\epsilon$-neighborhood'' of a given data point, we can set $\epsilon$ between [1, $\sqrt{2}$] which includes only the six first nearest neighbors. To expand the $\epsilon$-neighborhood, we can choose the value of $\epsilon$ in [$\sqrt{2}$, $\sqrt{3}$] to include the twelve second nearest neighbors, and so on. In the following visualization and discussion, we keep $\epsilon$ fixed at 1.7 so that only the 18 nearest neighbors (1st and 2nd) of a given data point are picked for the density calculation during local clustering.

The most critical adaptation to apply DBSCAN on our neutron dataset is to use the intensity as a measure of weight in calculating the second DBSCAN parameter –- \textit{minPts}. As mentioned above, the intensity of neutron scattering data is of physical significance and it is obvious that the diffuse patterns are as much spatially correlated as photometrically. To utilize both information, \textit{i.e.}, of the intensity and spatial location, we dictate DBSCAN to calculate \textit{minPts} with varying weights so that for each data point, its contribution to weighted-\textit{minPts} is proportional to its intensity. By doing so, the DBSCAN algorithm becomes very effective in de-noising and feature/boundary detection for neutron scattering data. For example, with a proper weighted-\textit{minPts}, one can detect both the Bragg peaks (sharp spots with a few high intensity points) and the diffuse scattering patterns (broad features with many low intensity points) and label them in different clusters provided sufficient spatial separations.

\begin{figure}[!h]
	\includegraphics[width=0.48\textwidth]{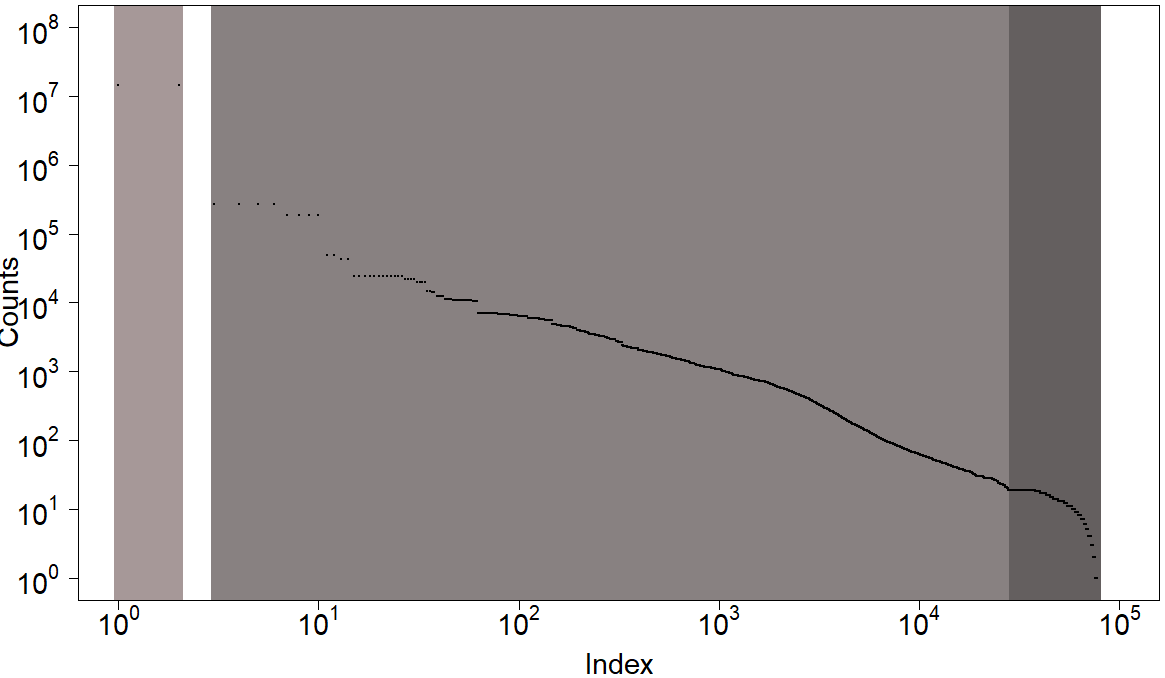}
	\caption{The number of data points as a function of the cluster index. The clusters identified by DBSCAN are sorted by the cluster size in the descending order. Three groups of clusters can be categorized as: a. index$\in$[1,2]; b. index$\in$[3,27865] and c. index$\in$[27866,Max. Index], here ``Max. Index'' is 77132.}
	\label{fig:Clusters}
\end{figure}
\begin{figure}[!h]
	\begin{center}
		\includegraphics[width=0.5\textwidth]{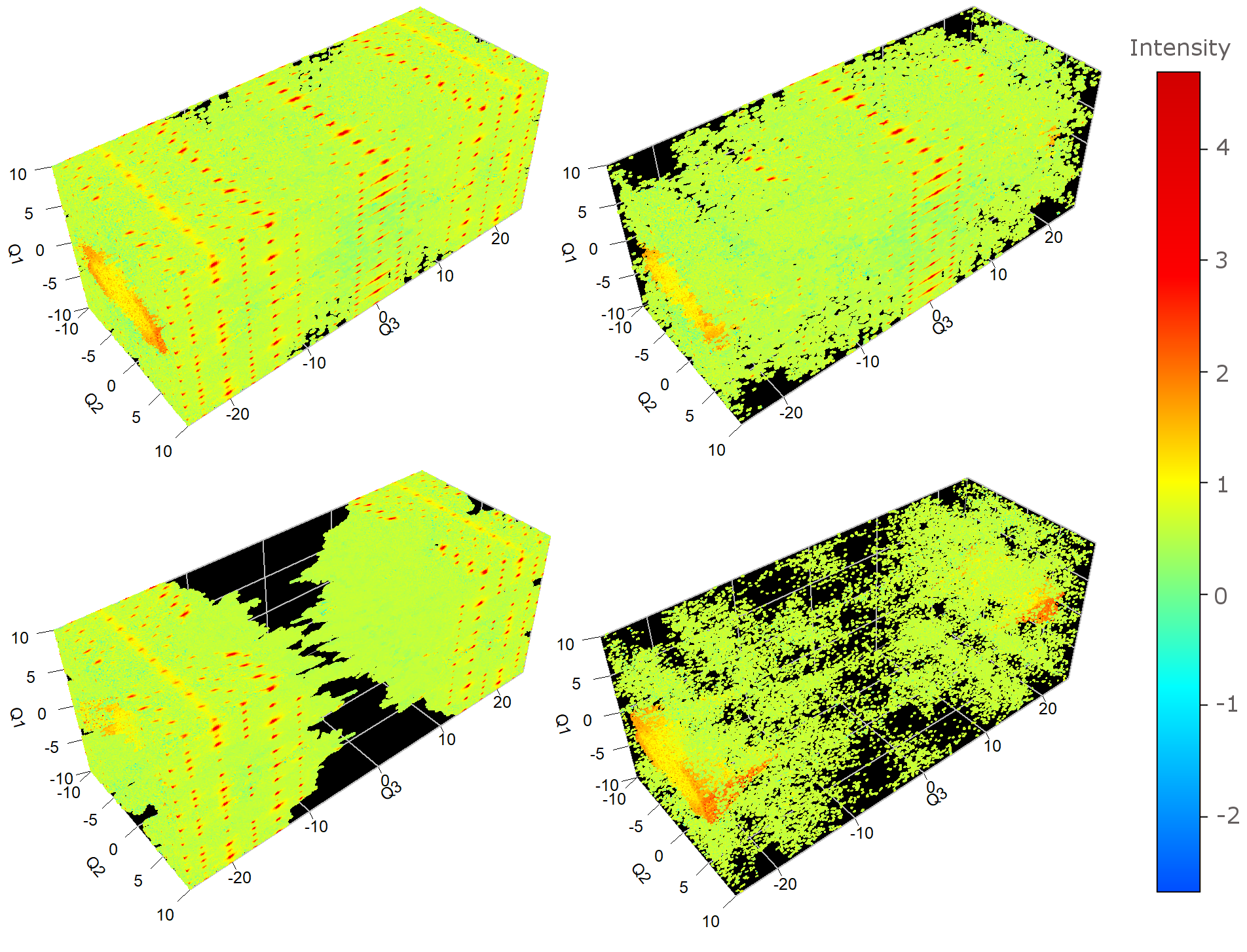}
		\caption{Reduced datasets with different selected cluster indices: 1) Top Left -- all clusters included; 2) Bottom Left -- clusters with indices [1 -- 2]; 3) Top Right -- clusters with indices [3 -- 27865]; and 4) Bottom Right -- clusters with indices [27866 -- Max. Index].}
		\label{fig:IndexSections}
	\end{center}
\end{figure}

\subsection{Clustering in 3D space}
\label{sec:3DClustering}

Specifically speaking, we carry out all DBSCAN clustering processes by using the ``DBSCAN'' function provided in the R-``dbscan'' package~\cite{b5}. As mentioned before, we use the intensity as a measure of weight in ``DBSCAN'' and take a certain value of intensity as a THRESHOLD so that any data point with its intensity less than the THRESHOLD will take a weight of 1, while data points with intensities great than the THRESHOLD acquire their weights from the ratio of their intensities to the THRESHOLD. In all our calculations, we choose THRESHOLD as $0.3\times$\textcolor{blue}{VMEDIAN}. With ``$\epsilon$'' fixed at 1.7, we tested a series of values for \textit{minPts} and finally adopted \textit{minPts}$=80$ for all results shown below. 

The found clusters are sorted by the number of points in individual clusters in the descending order. \autoref{fig:Clusters} shows the clustering result:  the number of data points as a function of the cluster index. From this figure, we can easily see that 1) clusters are automatically grouped by a general symmetry (e.g., by 2, 4, 8, etc.) which reflects the both the crystal symmetry and the symmetrization step during the original data reduction.); 2) three distinctive regions of grouped cluster indices exist as of index from: a. [1 -- 2]; b. [3 -- 27865] and c. [27866 -- Max. Index], and ``Max. Index'' is 77132. Within the index group ``a'', DBSCAN identified 2 clusters containing 28.8 millions data points in total (or 22.9\% out of 126 millions). While in group ``b'' and ``c'', the total numbers of data points are 7830686 (6.3\%) and 589314 (0.5\%) , respectively.

\begin{figure*}[ht]
	\begin{center}
		\parbox{\textwidth}{
			\parbox{0.5\textwidth}{
				\includegraphics[width=0.48\textwidth]{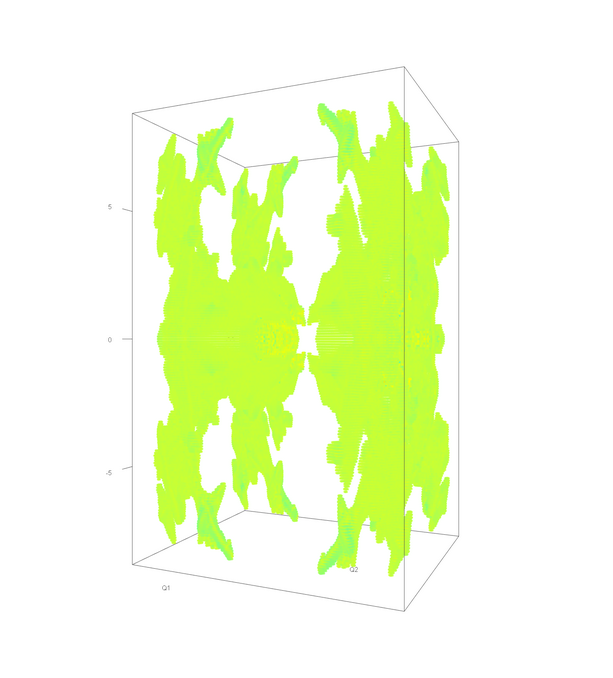}
			}
			\parbox{0.5\textwidth}{
				\includegraphics[width=0.48\textwidth]{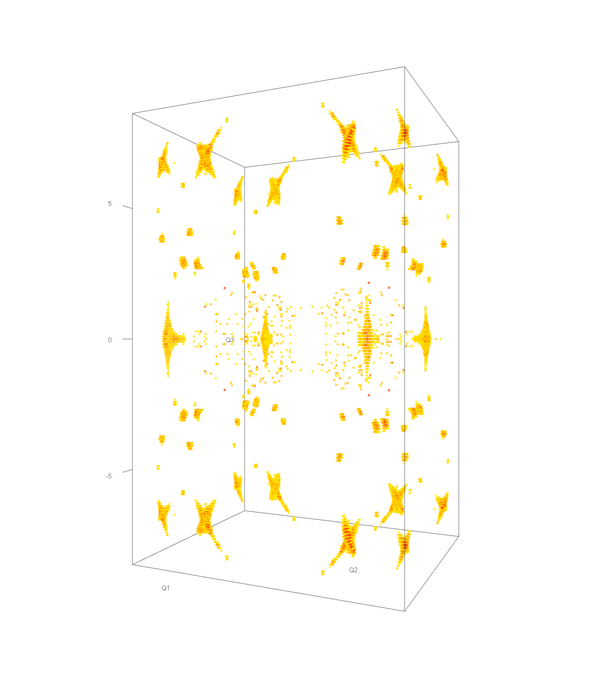}
			}
			\caption{3D plots of the center area illustrating the cluster group with indices from 7 to 10. LEFT: Intensity$\leq 8\times10^{-5}$; RIGHT: Intensity$> 8\times10^{-5}$.}
			\label{fig:GroupB}
		}
	\end{center}
\end{figure*}

\autoref{fig:IndexSections} shows the reduced dataset with selected clusters of interest. The used DBSCAN clustering parameters are: THRESHOLD = $0.3\times$\textcolor{blue}{VMEDIAN}, $\epsilon$ = 1.7 and \textit{minPts} = 80. Corresponding to the identified cluster indices regions, four panels in~\autoref{fig:IndexSections} are dedicated to: 1) Top Left -- all clusters included; 2) Bottom Left -- clusters with indices [1 -- 2]; 3) Top Right -- clusters with indices [3 -- 27865]; and 4) Bottom Right -- clusters with indices [27866 -- Max. Index]. It's obvious that DBSCAN differentiates the scattering signals by identifying the low signal-to-noise region as index group ``a'' (or ``Bottom Left'' panel) and high signal-to-noise region as index group ``b'' (``Top Right''). It's also interesting to see that in the low signal-to-noise region exists a non-localized ``wedge''-shaped feature with relatively higher intensities comparing to its surroundings. The two clusters in the index group ``a'' are actually enormous mixtures of highly entangled features of both Bragg peaks and diffuse scattering, and it becomes very challenging to further divide them into small meaningful clusters because of the low signal/noise ratio. Below, we will focus our investigation on the index groups ``b'' and ``c'' only.

\section{Feature Identification and Extraction}
\label{sec:IdentificationExtraction}

\subsection{Index group ``b''}
\label{sec:GroupB}

\begin{figure}[!h]
	\begin{center}
		\includegraphics[width=0.5\textwidth]{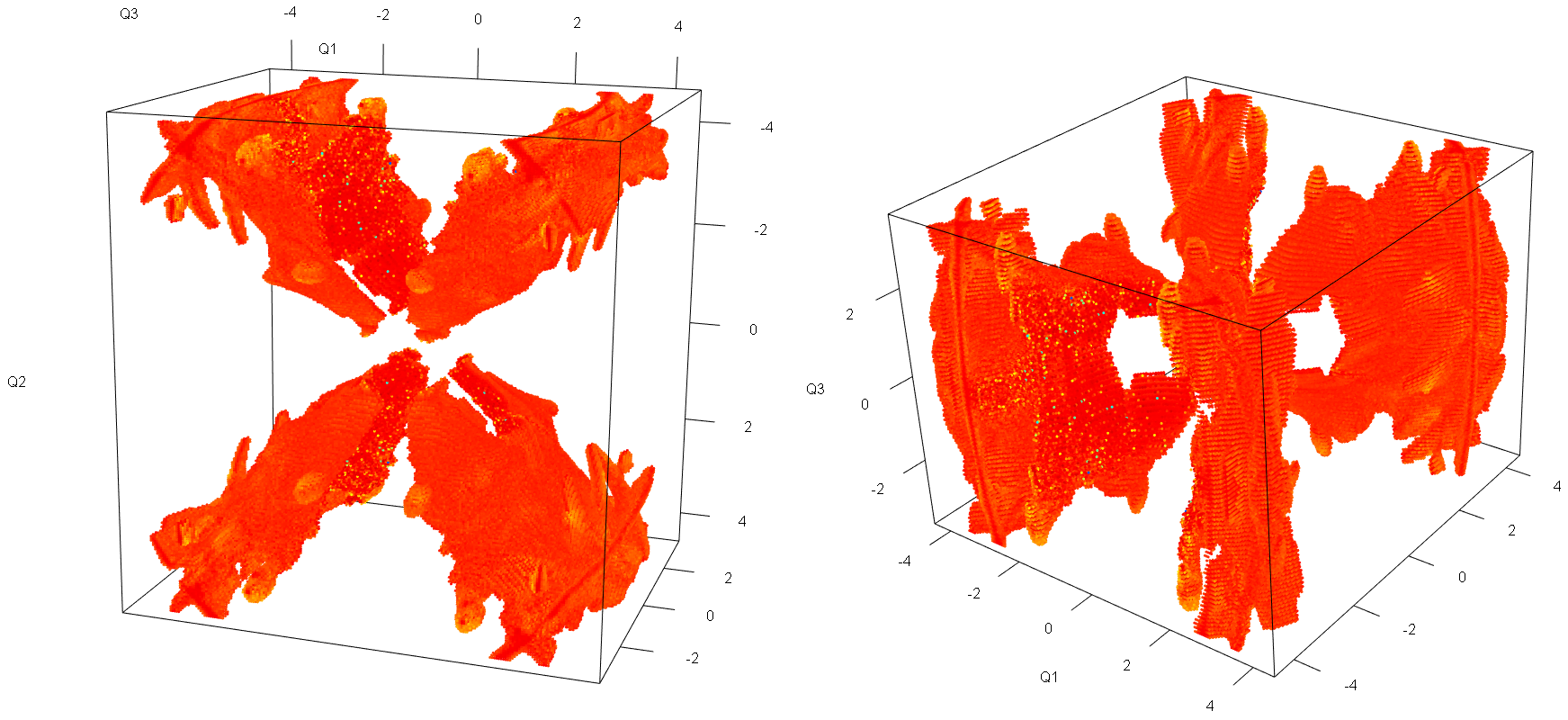}
		\caption{Close-up of the center region of cluster group C7-10, showing the ``double-cone'' structure with a different color key to make the feature stand out.}
		\label{fig:minPts80CloseUp}
	\end{center}
\end{figure}

\begin{figure*}[ht]
	\begin{center}
		\parbox{\textwidth}{
			\parbox{0.5\textwidth}{
				\includegraphics[width=0.48\textwidth]{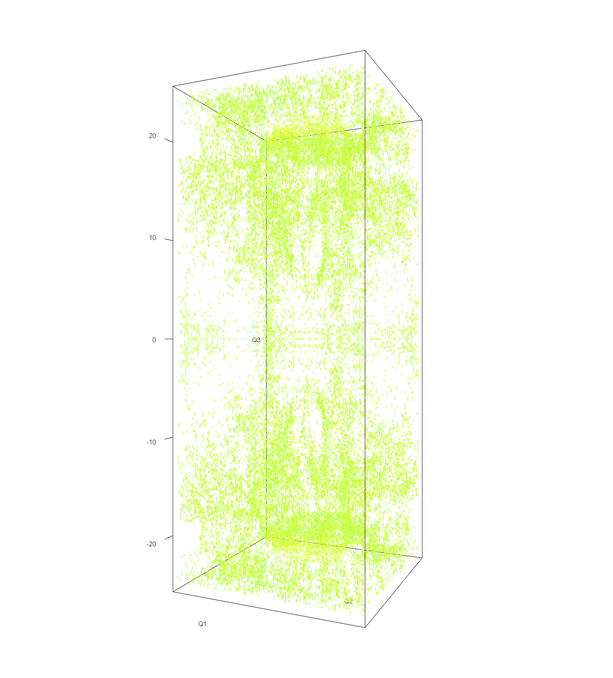}
			}
			\parbox{0.5\textwidth}{
				\includegraphics[width=0.48\textwidth]{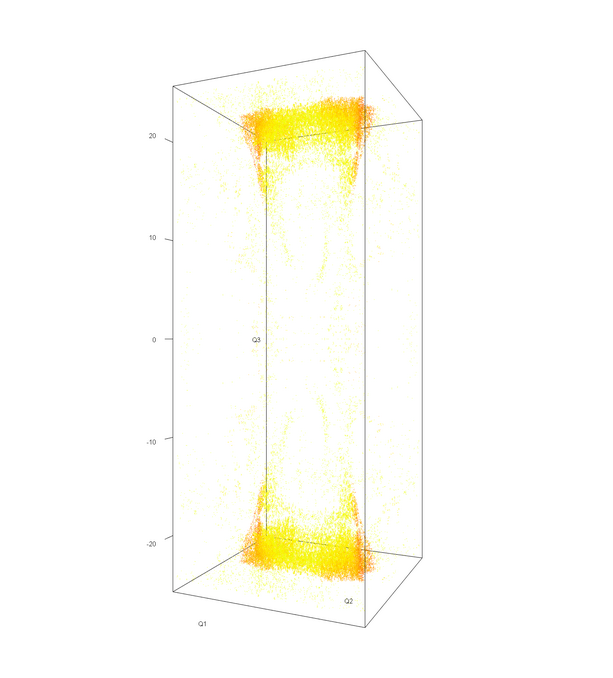}
			}
			\caption{3D plots of the remote-end area illustrating the cluster group with indices from 27866 to ``Max. Index''. LEFT: Intensity$\leq 4\times10^{-5}$; RIGHT: Intensity$> 4\times10^{-5}$.}
			\label{fig:GroupC}
		}
	\end{center}
\end{figure*}

\begin{figure}[ht]
	\begin{center}
		\parbox{0.5\textwidth}{
			\includegraphics[width=0.48\textwidth]{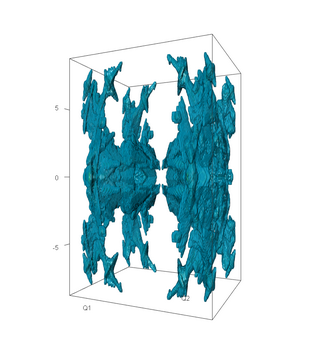}
		}
		\caption{An iso-surface redraw of the diffuse scattering patterns in~\autoref{fig:GroupB}.}
		\label{fig:IsoSurface}
	\end{center}
\end{figure}

As mentioned in~\autoref{sec:3DClustering} and shown in~\autoref{fig:GroupB} (also in~\autoref{fig:IsoSurface}), clusters identified by DBSCAN are automatically grouped by a general symmetry (e.g., by 2, 4, 8, etc.) reflecting in their total numbers of data points in each cluster. It is natural to inspect clusters by separating these clusters into small groups with symmetry-related clusters, e.g. clusters with indices from 3 to 6, or from 7 to 10, and so on, and then inspect each small group. Since there are almost 28 thousands of clusters in index group ``b'', it's impractical to go over every single one of them (or by their symmetry in total numbers of data points). Fortunately, we find that all the clusters up to cluster index ``154''represent Bragg peaks, except one small group of clusters with indices from 7 to 10 (hereafter, ``C7-10''). These four clusters represent the dominated diffuse scattering features in this dataset, which contains both Bragg peaks and the surrounded diffuse scattering signals.

For the purpose of better illustration, we plot the cluster group C7-10 with two different intensity filters. The left panel in~\autoref{fig:GroupB} shows C7-10 with a low-pass filter which includes all data points whose intensities $\leq 8\times10^{-5}$, while the right panel with a high-pass filter only accepting data points with intensities $> 8\times10^{-5}$. Note that the intensity values have been scaled by the ``1/THRESHOLD'' (for definition of THRESHOLD, refer to~\autoref{sec:3DClustering}). The most outstanding features we can identify in the diffuse scattering is that there exists a huge ``double-cone'' structure in the reciprocal space (spec. in regions defined by $Q_1\in[-4.5, 4.5]$, $Q_2\in[-4.5, 4.5]$ and $Q_3\in[-3.5, 3.5]$, as shown in~\autoref{fig:minPts80CloseUp}). The other significant broad feature is the long extending bars at the four corners of \autoref{fig:GroupB} along the axis of $Q_3$. On the other hand, the right panel in~\autoref{fig:GroupB} clearly shows many Bragg peaks with various characteristic shapes. Among them, three categories can be further identified as 1) ``X''-shaped tails from Bragg peaks which usually appear in regions with high $Q_1$, $Q_2$ and $Q_3$ values; 2) ``T''-shaped Bragg peaks which exist only along the plane defined by $Q_3=0$; and 3) irregular ``blobs'' whose locations are not strictly confined geometrically.

\subsection{Index group ``c''}
\label{sec:GroupC}

In~\autoref{fig:GroupC}, we show clusters in the index group ``c'' which interestingly stand out in a region with relatively lower signal-to-noise. Similar treatment in the intensity space is adopted as in~\autoref{sec:IdentificationExtraction} so that all clusters with 
index$\in$[27866, Max. Index] are plotted collectively in two panels after two filters (low-pass and high-pass) are applied.It's obvious that the unusual 
\begin{figure}[!h]
	\begin{center}
		\includegraphics[width=0.48\textwidth]{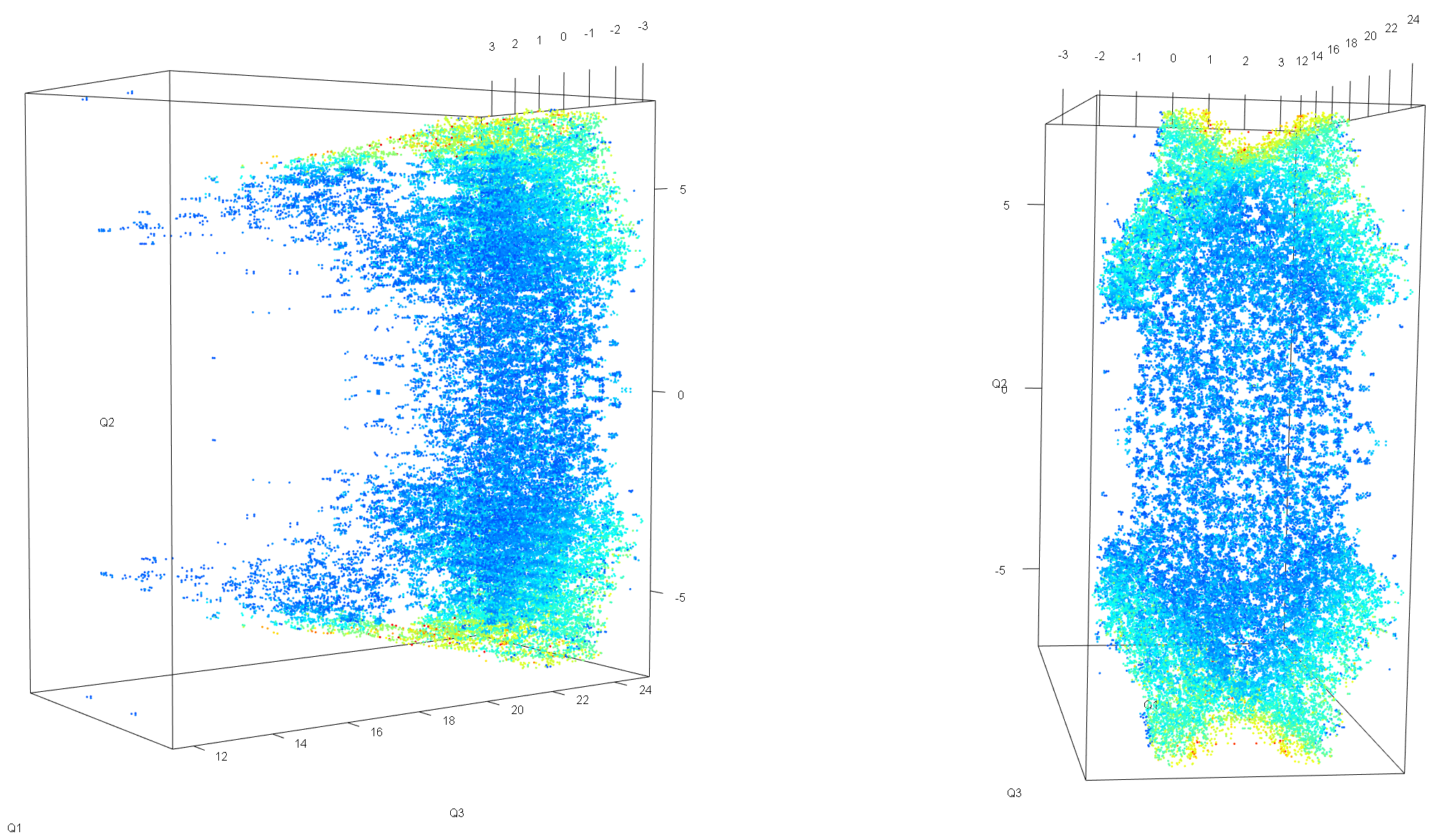}
		\caption{Close-ups of the remote-end region of index group ``c'', showing the ``broom-stick'' structure with a different color key to make the feature stand out.}
		\label{fig:TailCloseUp}
	\end{center}
\end{figure}
\begin{figure}[!h]
	\begin{center}
		\includegraphics[width=0.5\textwidth]{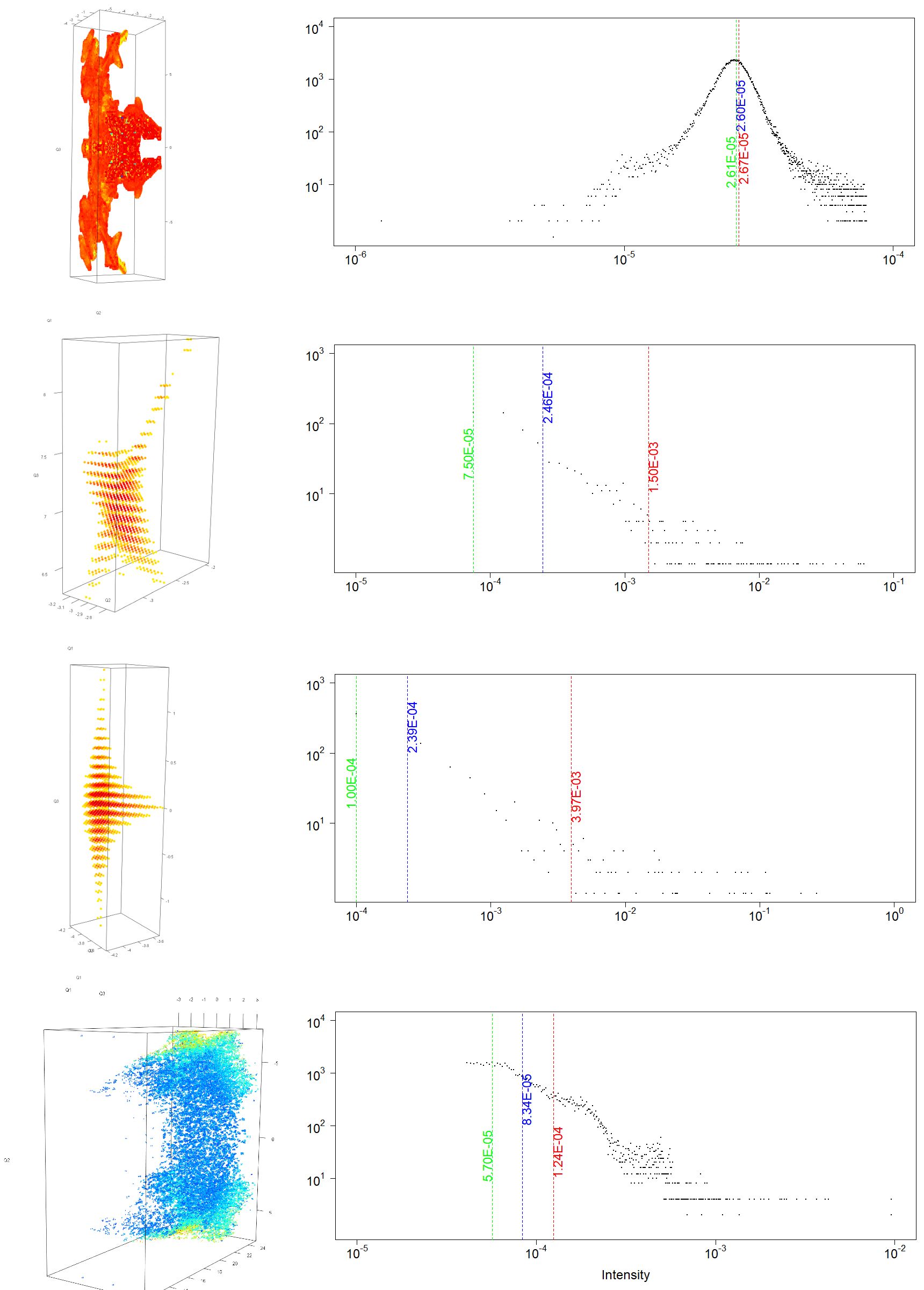}
		\caption{Scattering features characterized by both spatial and photometric distributions.}
		\label{fig:Characterization}
	\end{center}
\end{figure}
strong signals we are searching for almost exclusively appear in the right panel of ~\autoref{fig:GroupC}.

Further close-ups shown in~\autoref{fig:TailCloseUp} reveal that the feature
is confined in a small region defined by $Q_1\in[-4, 4]$, $Q_2\in[-7, 7]$ and $Q_3\in[10, 25]$. The overall shape of the feature resembles that of a broomstick while the tip at $Q_3\sim25$ apparently has a highly regular patterns in terms of spatial distribution. This feature is likely due to some artifacts during data reduction, however its decisive origin needs future investigation. 

\section{Characterization of Distinctive Features}
\label{sec:Characterization}

Once we identify and separate the desired features, we can characterize them by plotting a simple histogram in the intensity space and mark the statistically meaningful positions such as the mean(\textcolor{red}{VMEAN}) and median(\textcolor{blue}{VMEDIAN}) values, as well as the position (in intensity) of the maximum binned counts at \textcolor{green}{HMAX}. In~\autoref{fig:Characterization}, we pick all the significant features discussed in previous sections and illustrate both the spatial and photometric characteristics. From top to bottom we list: 1) a diffuse scattering pattern selected from a region defined by $Q_1\in[-4.5, 0]$, $Q_2\in[-4.5, 0]$ and $Q_3\in[-10, 10]$; 2) an ``X''-shaped Bragg peak confined in $Q_1\in[-4, -2]$, $Q_2\in[-3.5, -1.8]$ and $Q_3\in[6.2, 9]$; 3) a ``T''-shaped Bragg peak in $Q_1\in[-4.5, -3.5]$, $Q_2\in[-5, -3]$ and $Q_3\in[-2, 2]$; and 4) one of the ``broomstick'' features which is in $Q_1\in[-4, 4]$, $Q_2\in[-7, 7]$ and $Q_3\in[10, 25]$.

We only include the the ``X''- and ``T''-shaped tails from the Bragg peaks here and ignore these irregular-shaped Bragg peaks (see ~\autoref{fig:Characterization}), which apparently less interesting. 

\section{Acknowledgment}
\label{sec:Acknowledgment}

This manuscript has been authored by UT-Battelle, LLC under Contract No. DE-XXXX-00ORXXXXX with the U.S. Department of Energy. The United States Government retains and the publisher, by accepting the article for publication, acknowledges that the United States Government retains a non-exclusive, paid-up, irrevocable, world-wide license to publish or reproduce the published form of this manuscript, or allow others to do so, for United States Government purposes. The Department of Energy will provide public access to these results of federally sponsored research in accordance with the DOE Public Access Plan (\url{http://energy.gov/downloads/doe-public-access-plan}).


\begin{thebibliography}{00}
	
	\bibitem{b1} J. Etrillarda, M. Braden, A. Gukasov, U. Ammerahl and A. Revcolevschi. ``Structural aspects of the spin-ladder compound Sr$_{14}$Cu$_{24}$O$_{41}$''. Physica C, 2004, vol. 403, pp. 290-–296.
	
	\bibitem{b2} X. Chen, D. Bansal, S. Sullivan, D. L. Abernathy, A. A. Aczel, J. Zhou, O. Delaire and L. Shi. ``Weak coupling of pseudoacoustic phonons and magnon dynamics in the incommensurate spin-ladder compound Sr$_{14}$Cu$_{24}$O$_{41}$''. Phys. Rev. B, 2016, vol. 94, pp. 134309. 
	
	\bibitem{b3} O. Arnold, et al., ``Mantid — Data analysis and visualization package for neutron scattering and µ SR experiments'', Nuclear Instruments and Methods, 2014, vol. A764, pp. 156--166.
	
	\bibitem{b4} M. Ester, H.-P. Kriegel, J. Sander and X. Xu. ``A density-based algorithm for discovering clusters a density-based algorithm for discovering clusters in large spatial databases with noise''. In Proceedings of the Second International Conference on Knowledge Discovery and Data Mining (KDD'96), 1996, Evangelos Simoudis, Jiawei Han, and Usama Fayyad (Eds.). AAAI Press 226--231.
	
	\bibitem{b5} M. Hahsler, M. Piekenbrock, S. Arya and D. Mount, ``dbscan: Density Based Clustering of Applications with Noise (DBSCAN) and Related Algorithms''. \url{https://cran.r-project.org/web/packages/dbscan/index.html}
	
\end{thebibliography}
\end{document}